\documentclass{article}

\PassOptionsToPackage{numbers}{natbib}
\usepackage[preprint]{neurips_2026}

\usepackage[utf8]{inputenc} 
\usepackage[T1]{fontenc}    
\usepackage{hyperref}       
\usepackage{url}            
\usepackage{booktabs}       
\usepackage{amsfonts}       
\usepackage{nicefrac}       
\usepackage{microtype}      
\usepackage[table]{xcolor}  
\usepackage{colortbl}       
\usepackage{amsmath}        
\usepackage{enumitem}
\usepackage{graphicx}
\usepackage{ulem}
\usepackage{algorithm}
\usepackage{algpseudocode}
\usepackage{amssymb}
\usepackage{caption}
\usepackage{subcaption}
\usepackage{multirow}
\usepackage{array}
\usepackage{amsthm}         

\title{Learning Implicit Bias in Generative Spaces for Accelerating Protein Dynamics Emulation}

\author{%
  Kaihui Cheng\textsuperscript{1,2} \\
  \And
  Zhiqiang Cai\textsuperscript{1} \\
  \And
  Wenkai Xiang\textsuperscript{1} \\
  \And
  Zhihang Hu\textsuperscript{1} \\
  \AND
  Siyu Zhu\textsuperscript{1,2,3,$\dagger$} \\
  \And
  Tzuhsiung Yang\textsuperscript{1,$\dagger$} \\
  \And
  Yuan Qi\textsuperscript{1,2,$\dagger$} \\
  \AND
  \textsuperscript{1}Shanghai Academy of AI for Science \quad
  \textsuperscript{2}Fudan University \quad
  \textsuperscript{3}Shanghai Innovation Institute \\[3pt]
  \textsuperscript{$\dagger$}Corresponding authors \\[3pt]
  \texttt{siyuzhu@fudan.edu.cn} \quad
  \texttt{yangzixiong@sais.org.cn} \quad
  \texttt{qiyuan@fudan.edu.cn}
}

\begin{document}

\maketitle

\begin{abstract}
Generative emulators of protein dynamics produce plausible trajectories at a fraction of the cost of molecular dynamics, but they inherit their training distribution and tend to revisit known states rather than reach rare ones under long-horizon extrapolation. Inspired by classical enhanced sampling, we introduce an implicit, history-dependent bias in the generative space of a pretrained emulator. Specifically, a history-aware score estimator augments the frozen emulator with a distance-weighted bias that steers reverse-time sampling away from previously generated structures, regularized by an environment-support term. To preserve structural validity at long horizons, a score-based refinement step re-projects drifted samples onto the data manifold using the frozen emulator. Our experiments demonstrate that the method (i) raises diversity by $35\%$ on DynamicPDB-80; (ii) on $12$ zero-shot Fast-Folding proteins, the learned bias alone reaches the unbiased emulator's coverage up to ${\sim}15\times$ faster, and pairing it with refinement reaches the coverage up to ${\sim}37\times$ faster while covering ${\sim}3\times$ as many low-energy states.
\end{abstract}

\section{Introduction}
\label{sec:intro}

Generative emulators of protein dynamics reproduce short trajectories at a fraction of the cost of molecular dynamics (MD), yet their long-horizon rollouts tend to revisit known states rather than reach rare ones.
These rare states underlie biological processes including catalysis, transport, and folding~\cite{guo2016protein,bu2011proteins}, but sit behind high free-energy barriers that direct MD rarely crosses on routine timescales~\cite{henin2022enhanced,juraszek2012transition}.
Classical enhanced-sampling methods, including umbrella sampling~\cite{torrie1977nonphysical}, metadynamics~\cite{laio2002escaping,barducci2008well}, and replica exchange~\cite{sugita1999replica}, bias MD toward underexplored regions, but still incur substantial per-system simulation cost.
On the generative side, score-based diffusion~\cite{ho2020denoising,songdenoising,song2019generative,song2020score} extended with $SE(3)$-equivariant architectures~\cite{yim2023se,bose2024se,yim2023fast} has driven trajectory-level emulators that generate protein trajectories from sequence and structure conditioning~\cite{wu2023diffmd,arts2023two,jing2024generative,cheng20254d,fannddynafold}.
To encourage exploration in such emulators, recent work injects guidance through collective variables (CVs) or exploration objectives into the reverse process~\cite{nam2025enhancing,xie2026enhanced,richman2026unlocking}.

To accelerate long-horizon exploration, we introduce a history-dependent bias directly in the generative space of a pretrained protein dynamics emulator, leaving the emulator itself frozen.
A history-aware score estimator augments the frozen emulator's reverse-time score with a bias conditioned on the rollout history, such that the same noisy state yields different biased reverse scores under different histories and the bias adapts as the history bank updates, analogously to history-dependent biasing in molecular dynamics.
During training, the biased reverse step is regularized by an environment-support term that keeps the score on the emulator's data manifold.
At inference, a complementary score-based refinement step re-projects samples back onto this manifold through the frozen emulator, correcting drift that accumulates over long-horizon rollouts.
The left panel of Figure~\ref{fig:framework} illustrates the bias steering reverse-time sampling away from previously visited conformations.

Our main contributions are summarized as follows. Both components operate on a pretrained protein dynamics emulator that remains frozen throughout:
\begin{itemize}
    \item We propose a history-dependent bias in the generative space of a pretrained protein dynamics emulator. A history-aware score estimator augments the reverse-time score with a distance-weighted bias conditioned on past rollout frames, regularized during training by an environment-support term that keeps the biased score on the emulator's data manifold.
    \item We introduce a score-based refinement step that re-projects samples onto the emulator's data manifold through a short forward-then-reverse diffusion. Applied at inference, it sustains structural validity under strong bias and on horizons well beyond the training timescale.
    \item We empirically validate the method at both short- and long-horizon scales: on DynamicPDB-80, the bias raises diversity (mDiv-mean) by $35\%$; on $12$ zero-shot Fast-Folding proteins, the bias alone reaches the unbiased emulator's coverage up to ${\sim}15\times$ faster, and pairing it with refinement reaches the coverage up to ${\sim}37\times$ faster while covering ${\sim}3\times$ as many low-energy states.
\end{itemize}

\section{Related Work}
\subsection{Generative Modeling of Protein Structure and Conformational Ensembles}
Generative modeling of protein structure has progressed from single-state prediction to richer conformational distributions, yet the targets remain state-level rather than trajectory-level.
Deep learning methods such as AlphaFold2~\cite{jumper2021highly2}, RoseTTAFold~\cite{baek2021accurate}, and ESMFold~\cite{lin2022language} predict protein structures from sequence with high accuracy. Beyond deterministic prediction, generative approaches sample over conformational distributions~\cite{audagnotto2022machine,lundstr2str,jing2024alphafold,zheng2024predicting}, often parameterized by score-based diffusion~\cite{songdenoising,song2020score} on rigid-frame and $SE(3)$-equivariant representations~\cite{yim2023se,yim2023fast,bose2024se}. However, these methods generate conformations independently rather than as time-ordered trajectories.

\subsection{Generative Modeling of Protein Dynamics}
Generative emulators of protein dynamics fill this gap by learning the transitions between successive conformations directly from simulation trajectories. Among score-based trajectory models, DiffMD~\cite{wu2023diffmd} estimates score fields over molecular conformations; MDGen~\cite{jing2024generative} models protein motions relative to the input structure; DynaFold~\cite{fannddynafold} operates in a latent trajectory space; and AlphaFolding~\cite{cheng20254d} predicts future conformations from a reference structure and past frames. Force-field learners take a different route: DFF~\cite{arts2023two} learns a coarse-grained force field without force supervision, with a follow-up~\cite{plainer2025consistent} enforcing consistency with Langevin dynamics. While these methods improve short-horizon fidelity, repeated rollouts may concentrate in visited basins, reaching new conformations only after many extrapolation steps. Rather than training a new emulator, we freeze the pretrained one and learn a history-dependent bias in the generative space to accelerate long-horizon coverage.

\subsection{Enhanced Sampling and Controlled Generative Dynamics}
Enhanced sampling has long sought to bias exploration toward underrepresented regions. Classical approaches bias exploration through different mechanisms: umbrella sampling~\cite{torrie1977nonphysical} along CVs, replica exchange~\cite{sugita1999replica} across parallel temperatures, and metadynamics~\cite{laio2002escaping,barducci2008well}, which builds a history-dependent bias from past visits. Controlled diffusion adopts a similar mechanism in generative modeling, guiding reverse-time sampling by modifying the score or drift~\cite{songdenoising,dhariwal2021diffusion,ho2022classifier}. ConfDiff~\cite{wang2024protein} biases the frozen score with a force network learned from MD forces and energies, CV-guided diffusion~\cite{nam2025enhancing} steers reverse sampling along a chosen reaction coordinate, and recent diffusion-based formulations of enhanced sampling~\cite{xie2026enhanced,richman2026unlocking} embed exploration objectives into the reverse process. We instead build the bias from the rollout history itself, lifting the history-dependent biasing principle of metadynamics into the generative space of a frozen $SE(3)$ emulator, with an environment-support term preventing the learned bias from drifting off the emulator's manifold during training.

\begin{figure}[t]
    \centering
    \includegraphics[width=1.0\textwidth]{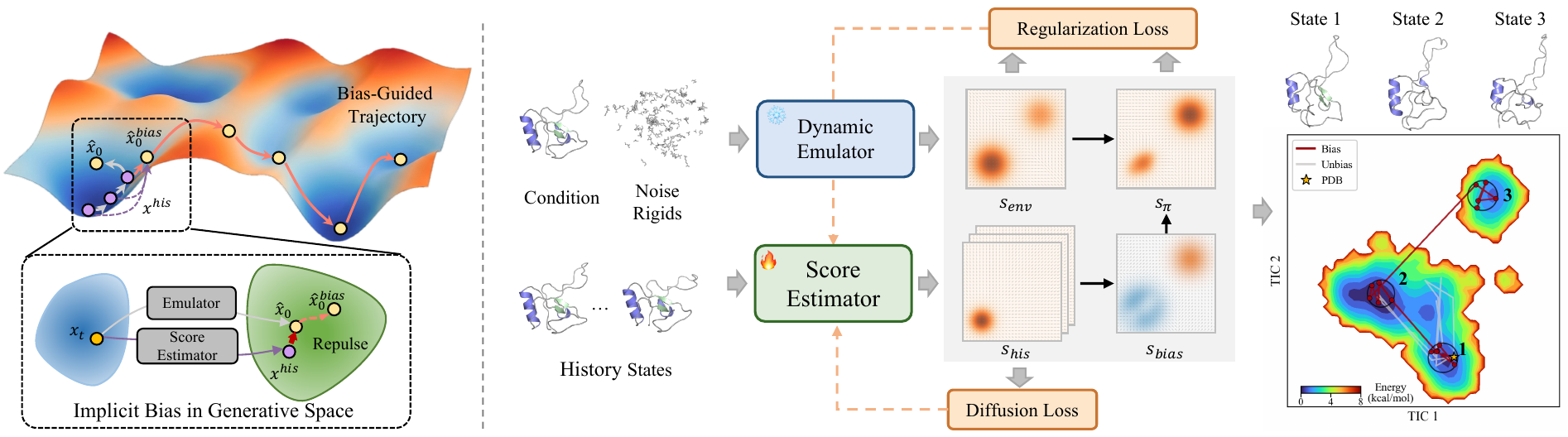}
    \caption{\textbf{Illustration of the proposed framework.}
    \textbf{Left:} concept. A history-aware score estimator adds a history-dependent bias to the reverse score; the generated structure $\hat{x}_0^{\mathrm{bias}}$ is shifted away from the trajectory history $x^{\mathrm{his}}$.
    \textbf{Middle:} architecture. A frozen emulator produces the reverse score $s_{\mathrm{emu}}$; the estimator produces history-aware scores $s_{\mathrm{his}}$, aggregated into a bias $s_{\mathrm{bias}}$ that is added to $s_{\mathrm{emu}}$ to form the biased reverse score $s_\pi$. Color overlays on the score fields encode the bias direction: dark red marks regions of higher biased density, while dark blue marks regions suppressed (penalized by the history). Training combines a diffusion loss with an environment-support regularizer.
    \textbf{Right:} the biased rollout (red) visits three metastable states on the TICA free-energy surface, while the unbiased rollout (gray) stays near the starting structure (yellow star).}
    \label{fig:framework}
\end{figure}

\section{Preliminaries}
\label{sec:preliminaries}

\subsection{Problem Formulation}
\label{subsec:problem_formulation}
We formulate long-horizon protein dynamics generation as an autoregressive trajectory problem. For clarity of exposition, we adopt a single-frame extrapolation setting in which each step generates one new structure conditioned on past frames; repeating this for $K$ steps produces a trajectory $\tau = (\mathbf{x}^{0}, \mathbf{x}^{1}, \dots, \mathbf{x}^{K-1})$ with $\mathbf{x}^{k} \in SE(3)^N$, where $k$ indexes frames along the trajectory and $N$ is the number of residues. Each step conditions on a residue sequence feature $\mathbf{c}$ shared across all frames and on a history bank $\mathcal{H} = \{\mathbf{h}^{0}, \mathbf{h}^{1}, \dots, \mathbf{h}^{H-1}\}$ of $H$ previously generated clean structures.
The frame $\mathbf{x}^{k}$ is sampled by a reverse diffusion process evolving from $t=1$ to $t=0$, with $\mathbf{x}_t^{k}$ denoting the noisy state at diffusion time $t$ and $\mathbf{x}_1^{k}$ drawn from the diffusion prior~\cite{yim2023se}. The frozen emulator parameterizes this process through a score $\mathbf{s}_{\mathrm{emu}}^\theta$ and a one-step denoising kernel $\mathbb{K}_{\mathrm{emu}}(\mathbf{x}_{t-\Delta t}^{k} \mid \mathbf{x}_t^{k},\, \mathbf{s}_{\mathrm{emu}}^\theta)$; our goal is to learn a history-aware score estimator that biases this reverse process using the rollout history.

\subsection{Score-Based Diffusion on $SE(3)$}
We build the emulator on the $SE(3)$ diffusion framework of~\cite{yim2023se}. Following AlphaFold2~\cite{jumper2021highly2}, each residue is represented by a rigid frame in $SE(3)$, and a protein backbone of $N$ residues lies in $SE(3)^N$. We decompose the noisy state as $\mathbf{x}_t = (\boldsymbol{\omega}_t, \boldsymbol{\nu}_t)$, where $\boldsymbol{\omega}_t \in \mathfrak{so}(3)$ is the rotation vector parameterizing an element of $SO(3)$ and $\boldsymbol{\nu}_t \in \mathbb{R}^3$ is the translation.
For the translational component on $\mathbb{R}^3$, the forward perturbation kernel is Gaussian, $q(\boldsymbol{\nu}_t \mid \boldsymbol{\nu}_0) = \mathcal{N}(\boldsymbol{\nu}_t; \alpha_t \boldsymbol{\nu}_0, \sigma_t^2 \mathbf{I})$, with score $\mathbf{s}_{\mathrm{trans}}(\boldsymbol{\nu}_t, t; \boldsymbol{\nu}_0) = \nabla_{\boldsymbol{\nu}_t} \log q(\boldsymbol{\nu}_t \mid \boldsymbol{\nu}_0)$.
For the rotational component on $SO(3)$, the forward perturbation kernel is the IGSO(3) density $f_t(\boldsymbol{\omega}_t \mid \boldsymbol{\omega}_0)$ conditional on the clean rotation $\boldsymbol{\omega}_0$~\cite{yim2023se}, with score $\mathbf{s}_{\mathrm{rot}}(\boldsymbol{\omega}_t, t; \boldsymbol{\omega}_0) = \nabla_{\boldsymbol{\omega}_t} \log f_t(\boldsymbol{\omega}_t \mid \boldsymbol{\omega}_0)$.

The emulator predicts a denoised rigid structure $\hat{\mathbf{x}}_0=(\hat{\boldsymbol{\omega}}_0, \hat{\boldsymbol{\nu}}_0)$ from $\mathbf{x}_t$, which we substitute as the clean state in the per-component score formulas above. The combined reverse-step score is $\mathbf{s}(\mathbf{x}_t, t) = (\mathbf{s}_{\mathrm{rot}}(\boldsymbol{\omega}_t, t; \hat{\boldsymbol{\omega}}_0),\ \mathbf{s}_{\mathrm{trans}}(\boldsymbol{\nu}_t, t; \hat{\boldsymbol{\nu}}_0))$, and we denote by $\mathbb{K}(\mathbf{x}_{t-\Delta t} \mid \mathbf{x}_t, \mathbf{s})$ the one-step reverse transition kernel obtained by an Euler--Maruyama update under score field $\mathbf{s}$.

\section{Methodology}
\label{sec:methodology}

As illustrated in Figure~\ref{fig:framework}, our method keeps the emulator introduced in Section~\ref{sec:preliminaries} frozen and trains a history-aware score estimator that augments the emulator's reverse score with a history-dependent bias. Because the bias is conditioned on the rollout history $\mathcal{H}$, the same noisy state $\mathbf{x}_t$ yields different biased scores under different histories, and the bias adapts as the history bank updates. For long-horizon rollouts, a complementary score-based refinement step re-projects drifted samples onto the data manifold using the same frozen emulator score, sustaining per-frame structural validity.

\subsection{Implicit Bias in Generative Space}
\label{subsec:bias_score}

Although the frozen emulator's reverse score $\mathbf{s}_{\mathrm{emu}}^\theta$ can faithfully reproduce its training trajectories, those trajectories rarely sample the equilibrium distribution: crossing the high free-energy barriers behind rare states requires timescales beyond what direct MD can routinely access~\cite{henin2022enhanced,juraszek2012transition}. As a result, rollouts driven by this score tend to revisit known states rather than reach rare ones, limiting long-horizon exploration. To address this, we learn a history-aware score estimator that augments $\mathbf{s}_{\mathrm{emu}}^\theta$ with a history-dependent bias, yielding a biased reverse score conditioned on the history $\mathcal{H}$.

\paragraph{History-dependent bias score.} Let $\phi$ parameterize the history-aware score estimator. The estimator has $H$ parallel branches, one per structure in the history bank. Given the current noisy state $\mathbf{x}_t^{k}$, diffusion time $t$, sequence feature $\mathbf{c}$, and history $\mathcal{H}$, branch $j$ outputs a score $\mathbf{s}_{\mathrm{his},j}^{\phi}(\mathbf{x}_t^{k}, t, \mathcal{H}, \mathbf{c})$ trained (Section~\ref{subsec:training}) to denoise $\mathbf{x}_t^{k}$ toward the history structure $\mathbf{h}^{j}\in\mathcal{H}$. The bias aggregates these branch scores as a weighted sum of differences from the emulator score,
\begin{equation}
\mathbf{s}_{\mathrm{bias}}^{\phi}\!\left(\mathbf{x}_t^{k}, t, \mathcal{H}, \mathbf{c}\right)
=
\sum_{j=0}^{H-1} w_j\,
\Bigl(\mathbf{s}_{\mathrm{emu}}^\theta(\mathbf{x}_t^{k}, t) - \mathbf{s}_{\mathrm{his},j}^{\phi}(\mathbf{x}_t^{k}, t, \mathcal{H}, \mathbf{c})\Bigr).
\label{eq:effective_bias_score}
\end{equation}
By construction, $\mathbf{s}_{\mathrm{bias}}^\phi$ points away from directions that denoise toward previously visited conformations. The biased reverse score is the emulator score augmented with $\mathbf{s}_{\mathrm{bias}}^\phi$ at strength $\eta\ge 0$,
\begin{equation}
\mathbf{s}_{\pi}^{\phi}\!\left(\mathbf{x}_t^{k}, t, \mathcal{H}, \mathbf{c}\right)
=
\mathbf{s}_{\mathrm{emu}}^\theta(\mathbf{x}_t^{k}, t)
+ \eta\,\mathbf{s}_{\mathrm{bias}}^{\phi}(\mathbf{x}_t^{k}, t, \mathcal{H}, \mathbf{c}).
\label{eq:score_perturbation}
\end{equation}

\paragraph{Branch weights.} The weights $\{w_j\}$ concentrate the bias on branches whose denoised prediction is closest to the emulator's. We parameterize them as a softmax over Kabsch-aligned C$\alpha$ root-mean-square deviation (RMSD),
\begin{equation}
w_j = \frac{\exp(-d_j/\kappa)}{\sum_{l=0}^{H-1}\exp(-d_l/\kappa)} ,
\label{eq:history_branch_weight}
\end{equation}
where $d_j$ is the C$\alpha$ RMSD between the emulator's and branch $j$'s denoised structures (both Kabsch-aligned), and $\kappa>0$ is the softmax temperature.

\paragraph{Sampling with the biased score.} At inference, the estimator enters only through $\mathbf{s}_\pi^\phi$: we apply the same Euler--Maruyama reverse update as the unbiased emulator, with $\mathbf{s}_\pi^\phi$ in place of $\mathbf{s}_{\mathrm{emu}}^\theta$,
\begin{equation}
\mathbb{K}_{\pi}\!\left(\mathbf{x}_{t-\Delta t}^{k}\,\middle|\,\mathbf{x}_t^{k}, \mathcal{H}, t\right)
=
\mathbb{K}\!\left(\mathbf{x}_{t-\Delta t}^{k}\,\middle|\,\mathbf{x}_t^{k}, \mathbf{s}_{\pi}^{\phi}(\mathbf{x}_t^{k}, t, \mathcal{H}, \mathbf{c})\right),
\label{eq:guided_kernel}
\end{equation}
preserving the parametric form of the frozen reverse dynamics.

\subsection{Loss Function}
\label{subsec:training}

We train the history-aware score estimator with two complementary objectives: a per-branch diffusion loss that supervises each branch $\mathbf{s}_{\mathrm{his},j}^\phi$ to denoise toward its history target $\mathbf{h}^j$, and an environment-support regularizer that prevents the biased reverse kernel $\mathbb{K}_\pi$ from drifting off the frozen emulator's manifold.

\paragraph{Per-branch diffusion loss.} Each branch $\mathbf{s}_{\mathrm{his},j}^\phi$ is trained to match the reverse-time score that denoises the current noisy state toward its history structure $\mathbf{h}^{j}$. Given a clean training structure $\mathbf{x}^{k}$ and a sampled diffusion time $t$, we perturb $\mathbf{x}^{k}$ forward into $\mathbf{x}_t^{k}$. For each $\mathbf{h}^{j}\in\mathcal{H}$, the target $\mathbf{s}_{\mathrm{target}}(\mathbf{x}_t^{k}, \mathbf{h}^{j}, t)$ is the closed-form reverse score that denoises $\mathbf{x}_t^{k}$ toward $\mathbf{h}^{j}$, obtained by substituting $\mathbf{h}^{j}$ as the clean state in the per-component score formulas of Section~\ref{sec:preliminaries}. The loss is the per-branch denoising score-matching objective,
\begin{equation}
\mathcal{L}_{\mathrm{diff}}
=
\mathbb{E}_{t,\,\mathbf{x}_t^{k}}\!\left[
\frac{1}{H}\sum_{j=0}^{H-1}
\gamma(t)\,\bigl\|
\mathbf{s}_{\mathrm{his},j}^{\phi}(\mathbf{x}_t^{k}, t, \mathcal{H}, \mathbf{c}) - \mathbf{s}_{\mathrm{target}}(\mathbf{x}_t^{k}, \mathbf{h}^{j}, t)
\bigr\|^2
\right],
\label{eq:diff_loss}
\end{equation}
where $\gamma(t)$ is the time-dependent loss weight from~\cite{yim2023se}.

\paragraph{Environment-support regularizer.} The per-branch diffusion loss trains each branch individually but does not constrain the resulting biased reverse kernel. At inference, Eq.~\eqref{eq:guided_kernel} evaluates the frozen reverse kernel at the biased score, so the biased score must propose next states where $\mathbf{s}_{\mathrm{emu}}^\theta$ remains reliable; otherwise the kernel sees out-of-distribution states and sample quality degrades. To prevent this, we regularize the expected KL divergence between the biased and frozen reverse kernels:
\begin{equation}
\mathcal{L}_{\mathrm{support}} = \mathbb{E}_{k,t}\!\left[\mathrm{KL}\!\left(\mathbb{K}_{\pi}(\cdot\mid\mathbf{x}_t^{k}, \mathcal{H}, t)\,\bigg\|\,\mathbb{K}_{\mathrm{emu}}(\cdot\mid\mathbf{x}_t^{k}, \mathbf{s}_{\mathrm{emu}}^\theta(\mathbf{x}_t^{k}, t))\right)\right].
\label{eq:support_loss_kl}
\end{equation}
Both kernels share the Gaussian transition form of Eq.~\eqref{eq:guided_kernel} and differ only in their reverse-drift score, so the KL admits an unbiased single-sample Monte Carlo estimator: we draw a proposal $\tilde{\mathbf{x}}_{t-\Delta t}^{k} \sim \mathbb{K}_{\pi}(\cdot \mid \mathbf{x}_t^{k}, \mathcal{H}, t)$ and evaluate the log-density ratio
\begin{equation}
\mathcal{L}_{\mathrm{support}}^{k}
=
\log \mathbb{K}_{\pi}\!\left(\tilde{\mathbf{x}}_{t-\Delta t}^{k}\,\middle|\,\mathbf{x}_t^{k}, \mathcal{H}, t\right)
- \log \mathbb{K}_{\mathrm{emu}}\!\left(\tilde{\mathbf{x}}_{t-\Delta t}^{k}\,\middle|\,\mathbf{x}_t^{k}, \mathbf{s}_{\mathrm{emu}}^\theta(\mathbf{x}_t^{k}, t)\right),
\label{eq:support_loss_frame}
\end{equation}
averaging $\mathcal{L}_{\mathrm{support}}\approx\mathbb{E}_{k,t}[\mathcal{L}_{\mathrm{support}}^{k}]$ over training frames and diffusion times.

The total loss is
\begin{equation}
\mathcal{L}_{\mathrm{total}}
=
\mathcal{L}_{\mathrm{diff}}
+
\lambda_{\mathrm{support}}\,\mathcal{L}_{\mathrm{support}},
\label{eq:total_loss}
\end{equation}
where $\lambda_{\mathrm{support}}\ge 0$ balances the two objectives: $\mathcal{L}_{\mathrm{diff}}$ shapes each branch's score toward its history target $\mathbf{h}^j$, while $\mathcal{L}_{\mathrm{support}}$ keeps the biased score on the frozen emulator's manifold during training. Appendix~\ref{subsec:girsanov} discusses a continuous-time SDE perspective on $\mathcal{L}_{\mathrm{support}}$.

\subsection{Refinement for Long-Horizon Rollouts}
\label{sec:method:refinement}

Long-horizon biased rollouts gradually drift off the data manifold, degrading per-frame structural validity. Prior work mitigates this drift by augmenting the emulator's conditioning structures with noise during training~\cite{shoghi2026scalable}; we instead repair drifted frames at inference time, without fine-tuning the emulator, by reusing the frozen emulator score $\mathbf{s}_{\mathrm{emu}}^\theta$ as a refinement projection. Each generated frame $\mathbf{x}^{\mathrm{gen}}$ is forward-diffused to a small noise level $t_{\mathrm{ref}}$ and then denoised by $S$ reverse steps under $\mathbf{s}_{\mathrm{emu}}^\theta$: $\hat{\mathbf{x}}^{\mathrm{gen}} \sim \mathbb{K}^{S}(\cdot \mid q(\mathbf{x}_{t_{\mathrm{ref}}} \mid \mathbf{x}^{\mathrm{gen}}), \mathbf{s}_{\mathrm{emu}}^\theta)$. The refinement runs as a post-processing step over generated frames, leaving the rollout history $\mathcal{H}$ and the learned bias unchanged.

\section{Experiments}

\subsection{Experimental Setup}\label{sec:setup}
\paragraph{Dataset.} We construct training and evaluation sets from DynamicPDB~\cite{liu2024dynamic}, a corpus of over $5{,}300$ protein trajectories curated from the Protein Data Bank (PDB)~\cite{berman2002protein} and simulated with OpenMM~\cite{eastman17openmm} under the Amber ff14SB force field for $100$~ns each. To prevent information leakage, we cluster all proteins at $40\%$ sequence identity with MMseqs2~\cite{steinegger2017mmseqs2} and split the clusters $80{:}20$, yielding $4{,}315$ training trajectories for emulator pretraining. For the score estimator, conditioning on a history bank of $H$ structures multiplies the per-frame memory footprint, so we restrict training to the $3{,}578$ trajectories with at most $224$ residues. For evaluation, we randomly sample $80$ proteins of up to $284$ residues from the held-out clusters, hereafter \textbf{DynamicPDB-80}; this budget keeps full-trajectory rollout and TICA reconstruction tractable across all evaluated methods.
For long-horizon extrapolation beyond the training distribution, we additionally evaluate on $12$ \textbf{Fast-Folding} proteins simulated on the Anton supercomputer by D. E. Shaw Research~\cite{lindorff2011fast}; these proteins are held out from DynamicPDB and evaluated zero-shot, with no per-protein fine-tuning.

\paragraph{Training.} The emulator is trained from scratch for $700$ epochs on $8$ NVIDIA H800 GPUs with the joint SE(3) diffusion objective, then frozen. The score estimator is trained for $300$ epochs on the same hardware, conditioning on a history bank of $H{=}3$ past structures, with branch-weight temperature $\kappa{=}1.0$ and support-regularization weight $\lambda_{\mathrm{support}}{=}0.5$. As a transferability test across emulator backbones, we train an additional estimator on top of the frozen pretrained ConfRover~\cite{shen2025simultaneous} under the same $\lambda_{\mathrm{support}}$. Full architecture and optimization details are in Appendix~\ref{app:impl:training}.

\paragraph{Inference.} At rollout time, we sample the initial noisy backbone from the $SE(3)$ diffusion prior and condition on the PDB starting structure and the residue-sequence embedding from Evoformer~\cite{jumper2021highly2}. Each generated frame uses $100$ reverse steps, and each rollout consists of $E$ autoregressive extrapolation segments of $16$ frames each, with $E$ varying by benchmark. Unless otherwise noted, we use bias strength $\eta{=}0.05$ on DynamicPDB-80 and $\eta{=}0.1$ on Fast-Folding. Refinement (Section~\ref{sec:method:refinement}) uses $t_{\mathrm{ref}}{=}0.2$ and $S{=}20$ steps. Full rollout settings and baseline configurations are in Appendix~\ref{app:impl:inference}.

\paragraph{Evaluation.} We evaluate in two timescale regimes: a short-timescale regime on DynamicPDB-80 with $E{=}16$ extrapolation segments, and a long-timescale regime on Fast-Folding with $E\geq 640$. Refinement is applied only in the long-timescale regime.

\paragraph{Structural validity.} For each generated frame, we compute three validity rates and average them over the trajectory. \textbf{C$\alpha$-Level Validity (CA\%)} is the fraction of adjacent C$\alpha$--C$\alpha$ pairs whose distance is below $4.5~\text{\AA}$; \textbf{Peptide-Bond Validity (CN\%)} is the fraction of adjacent peptide-bond C--N pairs whose distance is below $2.0~\text{\AA}$; and \textbf{Steric Validity (Clash-Free)} is the fraction of non-neighboring atom pairs separated by more than $1.0~\text{\AA}$.
\paragraph{Coverage and diversity.} All metrics are computed against the MD reference in a two-dimensional TICA space. A frame is compliant at threshold $\tau$ when its CA, CN, and Clash-Free rates all satisfy $\geq\tau$. \textbf{Ensemble Diversity (Div)} is the mean pairwise $1-\mathrm{lDDT}_{\mathrm{C}\alpha}$ over the generated ensemble. To trace the validity--diversity trade-off, we restrict Div to compliant frames, denoted $\mathrm{Div}@\tau$, and define \textbf{mDiv-mean} as the average of $\mathrm{Div}@\tau$ over $\tau\in[0.90, 0.99]$ on a 10-point grid. \textbf{Low-Energy State Coverage (Cov)}~\cite{lewis2025scalable} is the fraction of low-energy states (TICA regions with free energy below $4~\mathrm{kcal/mol}$) visited by $\tau{=}0.9$-compliant frames. To isolate the exploration gain from the bias, we treat each architecture's $\eta{=}0$ rollout as its own reference and report \textbf{Time-to-Coverage (TTC)}: the smallest frame count at which the biased coverage first exceeds the $\eta{=}0$ coverage at the full frame budget $F$. \textbf{Speedup} is $\mathrm{TTC}_{\eta=0}/\mathrm{TTC}_{\eta}$, the factor by which the bias reaches the reference coverage faster. Cov and TTC require an MD reference long enough to visit all metastable states; we therefore report them only on the Fast-Folding benchmark (Section~\ref{sec:long_timescale}).

\subsection{Short-Timescale Trajectory Emulation}
\label{sec:short_timescale}

\begin{table*}[t]
    \centering
    \caption{\textbf{Structural quality and diversity on DynamicPDB-80.}
    Metric definitions follow Section~\ref{sec:setup}; Div@$\tau$ is
    diversity over frames passing the compliance threshold $\tau$
    ($\tau{=}0$ recovers the raw diversity over all frames), and mDiv-mean
    averages Div@$\tau$ over $\tau\in[0.90, 0.99]$. Values are means over
    80 proteins. Shaded rows add our implicit bias on top of a frozen
    baseline.}
    \label{tab:short_cross_method}
    \small
    \resizebox{\textwidth}{!}{%
    \begin{tabular}{llccccccc}
        \toprule
        & & \multicolumn{3}{c}{Structural validity}
        & \multicolumn{4}{c}{Diversity} \\
        \cmidrule(lr){3-5} \cmidrule(lr){6-9}
        \multicolumn{2}{c}{Method}
            & CA\% $\uparrow$ & CN\% $\uparrow$ & Clash-Free $\uparrow$
            & Div@.00 $\uparrow$ & Div@.90 $\uparrow$ & Div@.95 $\uparrow$ & mDiv-mean $\uparrow$ \\
        \midrule
        \multicolumn{2}{c}{MD (Oracle)} & 100.00\% & 100.00\% & 100.00\% & 0.135 & 0.135 & 0.135 & 0.135 \\
        \midrule
        \multirow{3}{*}{Conformation Generation}
        & BioEMU~\cite{lewis2025scalable}     & 99.99\%  & 99.97\%  & 99.99\%  & 0.234 & 0.234 & 0.234 & 0.234 \\
        & AlphaFlow~\cite{jing2024alphafold}  & 100.00\% & 100.00\% & 99.99\%  & 0.122 & 0.122 & 0.122 & 0.122 \\
        & Str2Str~\cite{lundstr2str}          & 100.00\% & 100.00\% & 100.00\% & 0.192 & 0.192 & 0.192 & 0.192 \\
        \midrule
        \multirow{5}{*}{Trajectory Generation}
        & MDGEN~\cite{jing2024generative}     & 84.52\%  & 68.51\%  & 97.57\%  & 0.353 & 0.193 & 0.171 & 0.170 \\
        \cmidrule(lr){2-9}
        & ConfRover~\cite{shen2025simultaneous} & 97.47\% & 96.89\% & 99.26\% & 0.182 & 0.182 & 0.181 & 0.179 \\
        & \cellcolor{black!10}\ +\,Implicit Bias
            & \cellcolor{black!10}96.91\% & \cellcolor{black!10}96.34\% & \cellcolor{black!10}99.37\%
            & \cellcolor{black!10}0.221
            & \cellcolor{black!10}\textbf{0.221} & \cellcolor{black!10}\textbf{0.222} & \cellcolor{black!10}\textbf{0.212} \\
        \cmidrule(lr){2-9}
        & Emulator
            & 99.82\% & 98.43\% & 99.51\%
            & 0.185 & 0.185 & 0.184 & 0.181 \\
        & \cellcolor{black!10}\ +\,Implicit Bias
            & \cellcolor{black!10}98.37\% & \cellcolor{black!10}95.82\% & \cellcolor{black!10}98.87\%
            & \cellcolor{black!10}0.313
            & \cellcolor{black!10}\textbf{0.289} & \cellcolor{black!10}\textbf{0.252} & \cellcolor{black!10}\textbf{0.245} \\
        \bottomrule
    \end{tabular}}
\end{table*}
We first verify that the bias preserves per-frame structural validity at modest rollout depth, a precondition for any longer-horizon claim. Following Section~\ref{sec:setup}, every method generates $F{=}256$ frames per protein on DynamicPDB-80. Baselines fall into two classes: IID conformation generation, which samples frames independently, and trajectory generation, which propagates frames autoregressively. Table~\ref{tab:short_cross_method} reports per-frame structural validity and diversity at increasing compliance thresholds $\tau$.

Adding the bias improves diversity within each trajectory architecture while preserving structural validity. On our frozen emulator, the bias raises $\mathrm{Div}@.00$ from $0.185$ to $0.313$ with a CA drop of $1.45$ percentage points, and the gains persist at stricter thresholds (mDiv-mean from $0.181$ to $0.245$). The same mechanism transfers to ConfRover~\cite{shen2025simultaneous}: adding the bias raises $\mathrm{Div}@.00$ from $0.182$ to $0.221$ with CA dropping only $0.56$ percentage points (from $97.47\%$ to $96.91\%$), confirming that the bias is not specific to our emulator backbone.

Among trajectory baselines, MDGEN reaches the highest raw diversity ($0.353$) but at low validity (CA $84.5\%$, CN $68.5\%$), reflecting cumulative error in its autoregressive rollouts; once a compliance threshold is applied, its mDiv-mean drops to $0.170$. IID conformation generators are flat across $\tau$ by construction, since every frame is sampled independently.

\begin{figure}[t]
    \centering
    \begin{subfigure}[t]{0.245\linewidth}
        \includegraphics[width=\linewidth]{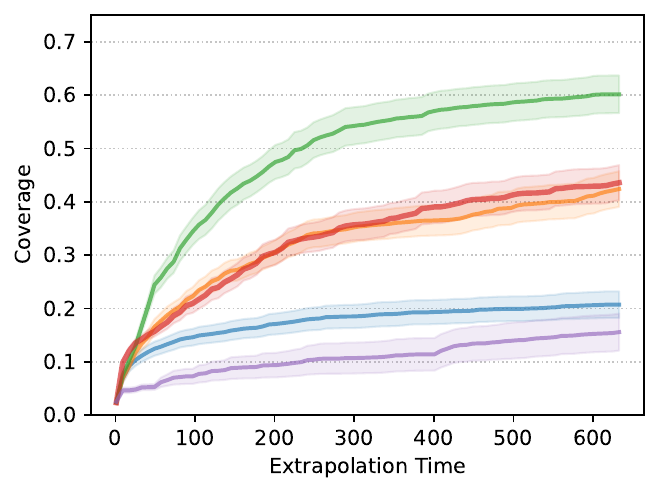}\\[1pt]
        \includegraphics[width=\linewidth]{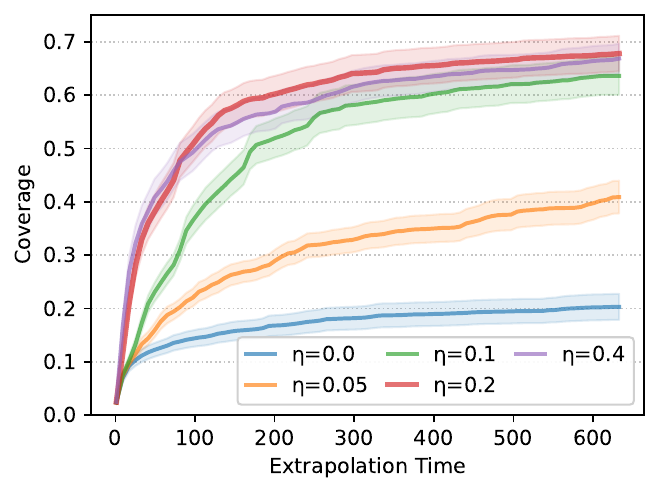}
        \caption{Coverage}\label{fig:ff_cov}
    \end{subfigure}\hfill
    \begin{subfigure}[t]{0.245\linewidth}
        \includegraphics[width=\linewidth]{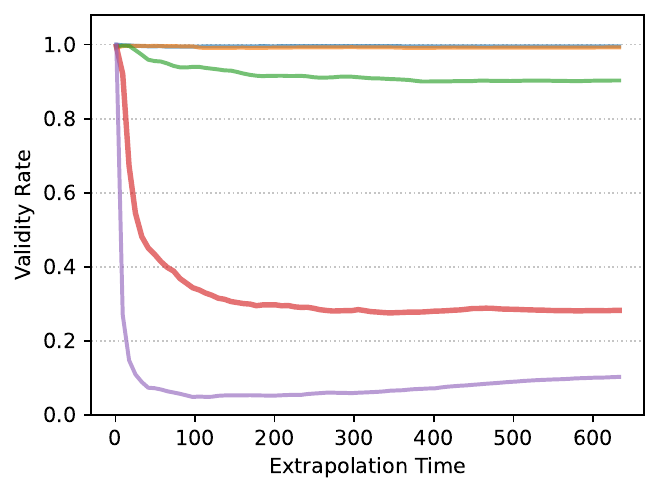}\\[1pt]
        \includegraphics[width=\linewidth]{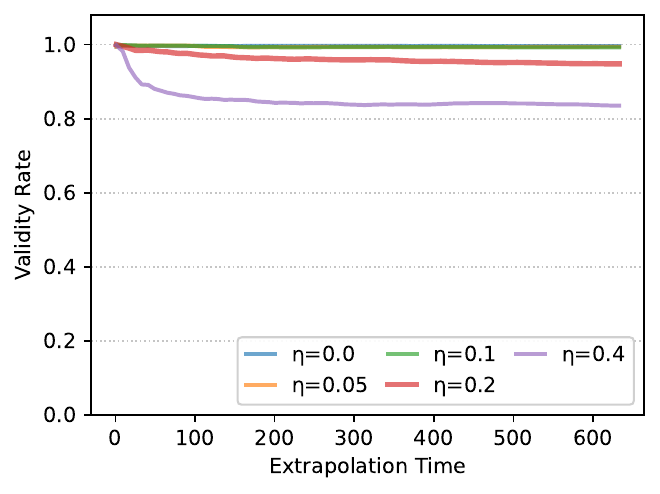}
        \caption{Validity Rate}\label{fig:ff_valid}
    \end{subfigure}\hfill
    \begin{subfigure}[t]{0.245\linewidth}
        \includegraphics[width=\linewidth]{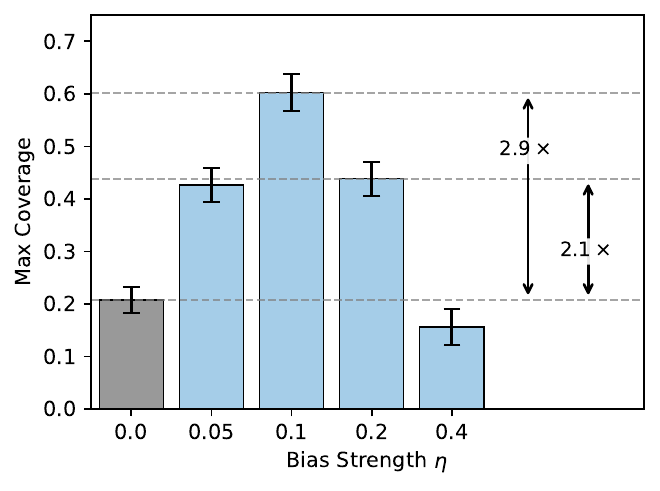}\\[1pt]
        \includegraphics[width=\linewidth]{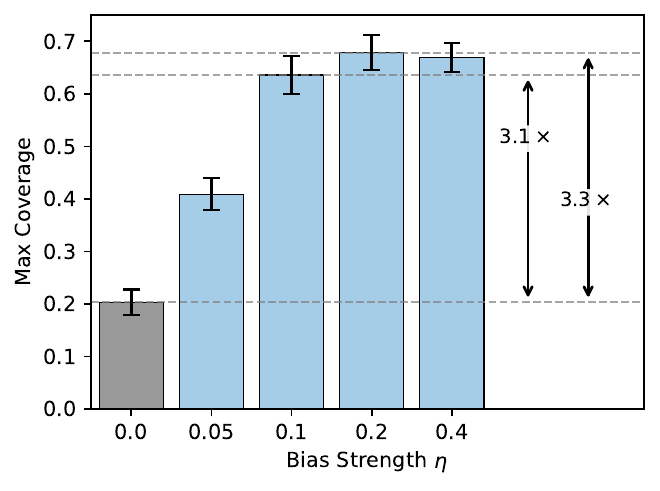}
        \caption{Max Coverage}\label{fig:ff_maxcov}
    \end{subfigure}\hfill
    \begin{subfigure}[t]{0.245\linewidth}
        \includegraphics[width=\linewidth]{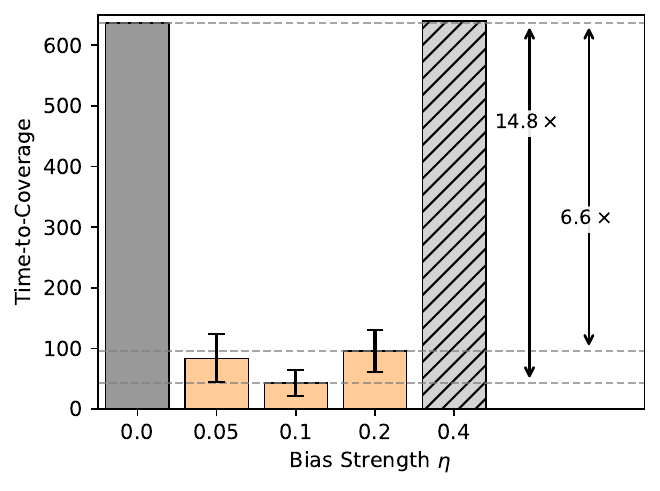}\\[1pt]
        \includegraphics[width=\linewidth]{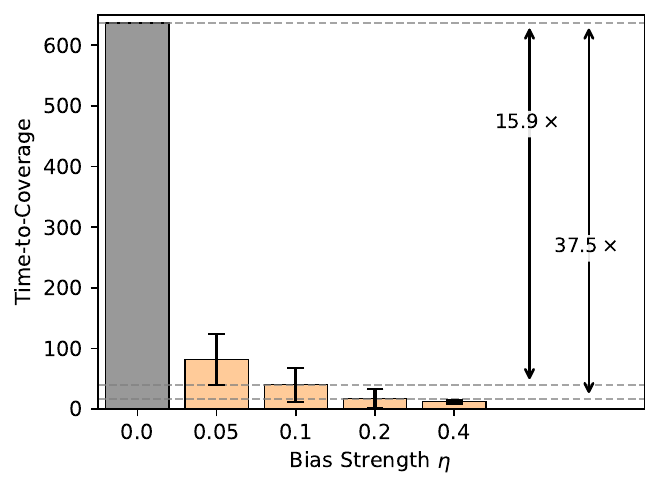}
        \caption{Time-to-Coverage}\label{fig:ff_ttc}
    \end{subfigure}
    \caption{\textbf{Refinement extends the usable range of bias strength $\eta$, recovering validity and low-energy state coverage that raw rollouts lose with long extrapolation.} Each panel stacks raw rollouts (top) above the same trajectories after refinement at $t_{\mathrm{ref}}{=}0.2$ (bottom): (\subref{fig:ff_cov}) coverage vs.\ extrapolation depth $E$, (\subref{fig:ff_valid}) cumulative validity rate, (\subref{fig:ff_maxcov}) max coverage vs.\ $\eta$, (\subref{fig:ff_ttc}) TTC vs.\ $\eta$, with hatched bars marking $\eta$ where TTC is never reached within the $E{=}640$ budget. Curves and bars show mean $\pm$SEM across the 12 Fast-Folding proteins, each averaged over 3 seeds.}
    \label{fig:fastfolding_long_timescale}
\end{figure}

\subsection{Long-Timescale Trajectory Extrapolation}
\label{sec:long_timescale}
We extend rollouts to microsecond horizons, far beyond the 100 ns training trajectories, and test whether the proposed method still improves conformational coverage without losing per-frame structural validity. The 12 Fast-Folding proteins from D. E. Shaw Research~\cite{lindorff2011fast} are held out from the DynamicPDB training set; we evaluate zero-shot, using a single checkpoint across all 12 systems rather than fine-tuning a separate model per protein as in leave-one-out cross-validation. We sweep bias strengths $\eta\in\{0, 0.05, 0.1, 0.2, 0.4\}$ across $3$ seeds, with each rollout spanning $E{=}640$ autoregressive extrapolation segments, totaling $10{,}240$ frames per protein. For raw biased rollouts, speedup peaks at $\eta{=}0.1$ and decays at higher $\eta$ as per-frame validity degrades, as shown in the top rows of Figure~\ref{fig:fastfolding_long_timescale}: $\eta{=}0.1$ yields $14.8\times$ over the $\eta{=}0$ baseline, $\eta{=}0.2$ yields $6.6\times$, and $\eta{=}0.4$ fails to reach baseline coverage within the $E{=}640$ budget.

To recover this lost speedup, we apply the refinement step of Section~\ref{sec:method:refinement} to the full biased rollouts. At $\eta{=}0.2$, refinement restores per-frame validity at $\tau{=}0.9$ from $28.3\%$ to $94.8\%$, raises max coverage to $3.3\times$ the $\eta{=}0$ baseline, and pushes speedup to $37.5\times$, as shown in the bottom rows of Figure~\ref{fig:fastfolding_long_timescale}. As a control, refining the unbiased $\eta{=}0$ rollout yields max coverage $0.203$ versus $0.208$ for the unrefined baseline, isolating the exploration gain to the bias. The full $\eta$ vs.\ refinement grid is in Appendix~\ref{app:bias_refine_control}. Figure~\ref{fig:ff_bias_coverage_examples} shows the biased rollout reaching metastable basins missed by the unbiased baseline on three representative proteins, raising low-energy state coverage from $0.29$ to $0.88$ on Protein G, $0.44$ to $1.00$ on WW domain, and $0.46$ to $0.73$ on $\alpha$3D.

\begin{figure}[t]
    \centering
    \begin{minipage}[t]{1.0\linewidth}
        \centering
        \includegraphics[width=\linewidth]{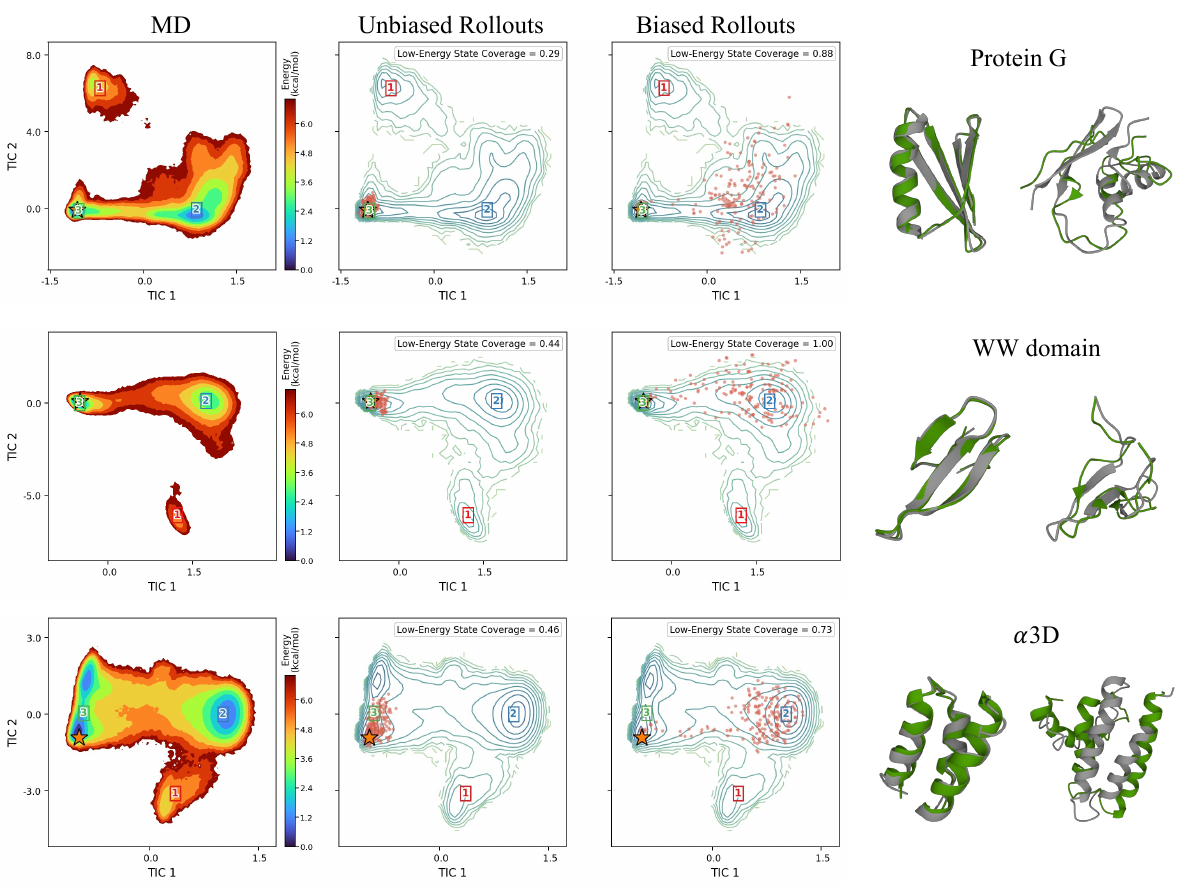}
        \captionof{figure}{\textbf{The bias expands low-energy state coverage on three representative proteins.} Column 1: MD reference free-energy surface with metastable basins labeled 1--3. Column 2: unbiased rollouts. Column 3: biased rollouts. Column 4: 3D structure overlay between MD reference (gray) and a representative biased frame (green). Red dots are TICA projections of generated frames; yellow star marks the input PDB structure.}
        \label{fig:ff_bias_coverage_examples}
    \end{minipage}
\end{figure}

\subsection{Ablation Studies}\label{sec:ablation}

\begin{figure}[t]
    \centering
    \includegraphics[width=1.0\linewidth]{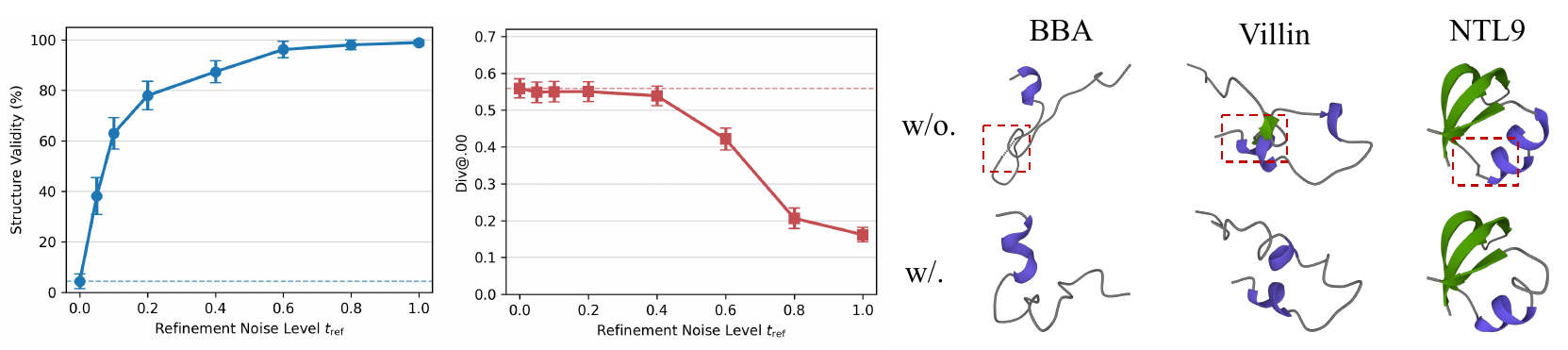}
    \caption{\textbf{Refinement noise level $t_{\mathrm{ref}}$ controls validity recovery while preserving diversity.} \textbf{Left}: structure validity at $\tau{=}0.9$ rises sharply with $t_{\mathrm{ref}}$ from the pre-refinement baseline (dashed). \textbf{Middle}: diversity (Div@.00) stays near the pre-refinement baseline (dashed) at low $t_{\mathrm{ref}}$, then drops as forward noising approaches the prior. \textbf{Right}: structure overlays on representative proteins, before (top) and after (bottom) refinement; broken local geometries (red boxes) are repaired.}
    \label{fig:ff_refinement_t_sweep}
\end{figure}

\paragraph{Refinement Noise Level.}
We sweep $t_{\mathrm{ref}}$ from $0.0$ to $1.0$ on the $50$ worst-validity segments per protein at $\eta{=}0.2$, where refinement is most needed. Figure~\ref{fig:ff_refinement_t_sweep} shows that the per-frame pass-rate at $\tau{=}0.9$ rises sharply with $t_{\mathrm{ref}}$ and saturates near $0.2$, climbing from $4.4\%$ to $77.9\%$ while preserving diversity; the right panels show that refinement repairs broken local geometries while keeping structures plausible. We adopt $t_{\mathrm{ref}}{=}0.2$ at this inflection, where validity has recovered while the forward marginal remains well short of the prior. At small $\eta$, refinement also does not degrade coverage (Appendix~\ref{app:bias_refine_control}).

\begin{figure}[t]
    \centering
    \begin{minipage}[t]{0.32\linewidth}
        \centering
        \includegraphics[width=\linewidth]{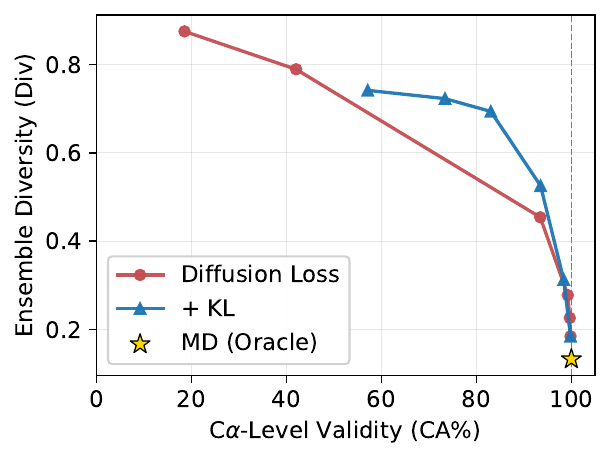}
        \captionof{figure}{\textbf{$\mathcal{L}_{\mathrm{support}}$ preserves validity across the $\eta$-sweep on DynamicPDB-80.} Validity--diversity curves trained with and without $\mathcal{L}_{\mathrm{support}}$. Yellow star: MD oracle.}
        \label{fig:ablation_loss_pareto}
    \end{minipage}
    \hfill
    \begin{minipage}[t]{0.32\linewidth}
        \centering
        \includegraphics[width=\linewidth]{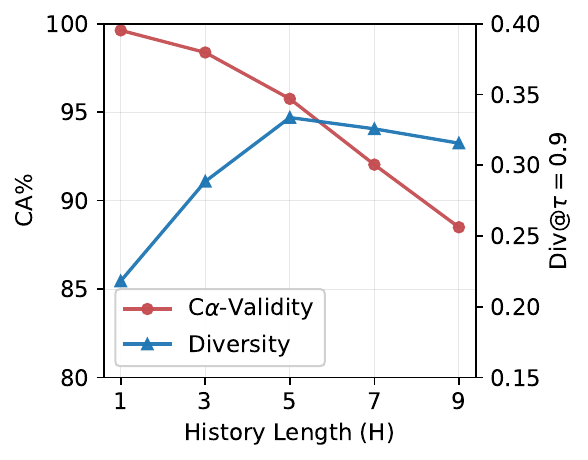}
        \captionof{figure}{\textbf{History length $H$ trades structural fidelity for diversity on DynamicPDB-80.} Increasing $H$ raises diversity while $\mathrm{C}\alpha$ validity decreases.}
        \label{fig:history_length_analysis}
    \end{minipage}
    \hfill
    \begin{minipage}[t]{0.32\linewidth}
        \centering
        \includegraphics[width=\linewidth]{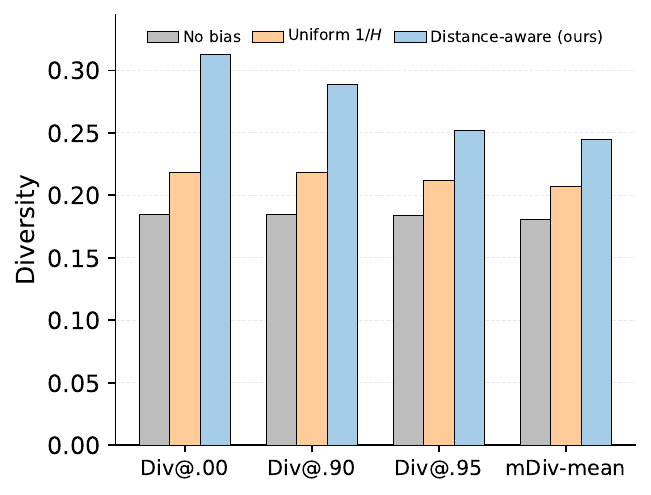}
        \captionof{figure}{\textbf{Weighting Strategies on DynamicPDB-80.} Distance-aware weighting outperforms uniform $1/H$ aggregation across compliance thresholds $\tau$.}
        \label{fig:branch_weighting_ablation}
    \end{minipage}
\end{figure}

\paragraph{Loss Function and Bias Strength.}
We sweep bias strength $\eta\in\{0.05,\ldots,1.0\}$ with and without the environment-support regularizer $\mathcal{L}_{\mathrm{support}}$. Figure~\ref{fig:ablation_loss_pareto} shows that without $\mathcal{L}_{\mathrm{support}}$, $\mathrm{C}\alpha$ validity drops sharply at high $\eta$, whereas adding $\mathcal{L}_{\mathrm{support}}$ preserves structural validity across the entire sweep and shifts the curve toward the upper-right, where both validity and diversity are higher.

\paragraph{History Length.}
We sweep the history length $H \in \{1, 3, 5, 7, 9\}$ to characterize the validity--diversity trade-off. Figure~\ref{fig:history_length_analysis} shows that diversity rises with $H$ and saturates while $\mathrm{C}\alpha$ validity drops. We adopt $H{=}3$, where diversity is near saturation and validity is still high.

\paragraph{Weighting Strategies.}
We compare our distance-aware weighting (Eq.~\ref{eq:history_branch_weight}) against a uniform $w_j{=}1/H$ aggregation at $\eta{=}0.05$, $H{=}3$, using the $\eta{=}0$ rollout as the no-bias reference. Figure~\ref{fig:branch_weighting_ablation} shows that uniform aggregation barely improves over the no-bias reference, whereas distance-aware weighting consistently captures most of the diversity gain across compliance thresholds.

\section{Discussion}
\label{sec:conclusion}

\paragraph{Limitations.}
Our experiments have validated history-dependent generative-space biasing for accelerated exploration of pretrained protein dynamics emulators. The current bias is unconditional: it explores the free-energy surface but cannot condition on a specific target endpoint. The reachable conformational space is bounded by the frozen emulator's training distribution: the bias redistributes density within this support, not beyond it. Finally, the base emulator does not target the equilibrium distribution, complicating post-hoc free-energy reconstruction via the multistate Bennett acceptance ratio (MBAR) or other schemes.

\paragraph{Opportunities.}
Beyond protein dynamics, history-dependent biasing in generative space may apply to long-horizon sampling in other pretrained physical-system generators. Goal-conditioned biasing, in which the score estimator is augmented with target-conformation inputs, would lift the unconditional limitation above by converting exploration bias into a controllable sampler for targeted conformational change. The framework can extend to multi-chain complexes and nucleic acid dynamics with appropriately retrained emulators. Uncertainty-aware sampling could prioritize the bias toward conformations where the emulator's score is most uncertain, allocating compute to under-explored regions of the free-energy surface. Finally, integrating the post-hoc Girsanov reweighting into a training-time objective could close the loop between biased rollouts and free-energy estimates.

\paragraph{Broader impacts.}
Accelerated exploration of protein conformational landscapes supports structural biology and drug discovery, including cryptic-pocket detection and allosteric-state characterization. The released models inherit the dual-use profile of public protein-structure generators and introduce no new high-risk capabilities; we release them under a permissive license without additional safeguards.

{\small
\bibliographystyle{plain}
\bibliography{egbib}
}

\appendix

\section{Implementation Details}
\label{app:impl}

\subsection{Training}
\label{app:impl:training}

\paragraph{Emulator backbone.}
The emulator is a joint $SE(3)$ frame diffusion network with sequence and trajectory-context conditioning. Single-residue and pair tracks have node embedding dim $384$ and edge embedding dim $128$, with $4$ Invariant Point Attention (IPA) blocks~\cite{jumper2021highly2}; each IPA block uses $8$ heads, $8$ query/key points, $12$ value points, and hidden dim $256$. Amino-acid type and residue-index features are projected to $32$-dim embeddings before the trunk.

\paragraph{Emulator training.}
We train from scratch on the DynamicPDB training partition with batch size $4$ per GPU on $8$ NVIDIA H800 GPUs, giving $32$ samples per optimizer step. Optimization uses Adam with AMSGrad at learning rate $10^{-4}$ with no warmup, no LR schedule. The objective is a sum of translation, rotation, backbone-atom, distance-matrix, and torsion losses with weights $1.0$, plus an auxiliary loss with weight $0.25$, following the joint $SE(3)$ formulation of~\cite{yim2023se}. Each training window contains $16$ frames, with the underlying MD trajectories subsampled at stride $40$. We train the emulator for $700$ epochs. The pretrained emulator is then frozen for the rest of the pipeline.

\paragraph{Score estimator.}
The history-aware score estimator shares the IPA backbone with the emulator (same dims, $4$ blocks, same head/point counts) but is a fully separate module trained independently. It conditions on a history bank of $H{=}3$ structures via a single shared trunk, where each history frame is folded into the pair track through the same self-conditioning pathway used by the emulator.

\paragraph{Score estimator training.}
We train the estimator on the same DynamicPDB training partition with batch size $1$, frame length $16$, history bank $H{=}3$, trajectory-segment stride $40$, and maximum residue length $224$. Optimization uses AdamW at learning rate $10^{-4}$ with weight decay $10^{-6}$ for $300$ epochs on $8$ NVIDIA H800 GPUs in fully synchronous data-parallel mode, with mixed precision off. The emulator backbone is frozen, so only the estimator weights receive gradients. The objective combines the per-branch diffusion loss of Eq.~\eqref{eq:diff_loss} with the environment-support regularizer of Eq.~\eqref{eq:support_loss_kl} at weight $\lambda_{\mathrm{support}}{=}0.5$, evaluated by the single-sample MC estimator of Eq.~\eqref{eq:support_loss_frame}. The transferability variant on the frozen pretrained ConfRover~\cite{shen2025simultaneous} backbone uses the same implementation.

\subsection{Inference and Baselines}
\label{app:impl:inference}

\paragraph{Rollout.}
At inference time we sample the initial noisy backbone from the joint $SE(3)$ diffusion prior used during training: translations follow a VP-SDE schedule with $\beta_{\min}{=}0.1$, $\beta_{\max}{=}20.0$ and a coordinate scaling factor of $0.1$, while rotations follow an IGSO(3) schedule with $\sigma\in[0.1,\,1.5]$ on a logarithmic grid of $1000$ points in both $\omega$ and $\sigma$. Each generated frame uses $T{=}100$ uniform Euler--Maruyama reverse steps from $t{=}1$ to $t_{\min}{=}0.01$, and each rollout produces $E$ autoregressive extrapolation segments of frame length $16$. Sequence conditioning uses MSA-based Evoformer~\cite{jumper2021highly2} embeddings, feeding the same $384$-dim node and $128$-dim pair tracks used at training. The bias strength is $\eta{=}0.05$ on DynamicPDB-80 and $\eta{=}0.1$ on Fast-Folding. The branch weighting uses the Kabsch-aligned RMSD with temperature $\kappa{=}1.0$. Refinement (Section~\ref{sec:method:refinement}) applies a partial forward noising to $t_{\mathrm{ref}}{=}0.2$ followed by $S{=}20$ environment-only reverse steps.
\paragraph{Baselines.}
For protein ensemble generation, we evaluate three baselines from publicly released pretrained checkpoints, each producing $256$ ensemble structures:
\begin{itemize}
    \item \textbf{AlphaFlow}~\cite{jing2024alphafold}: template conditioning enabled, $10$ denoising steps (released default).
    \item \textbf{BioEmu}~\cite{lewis2025scalable}: v1.1, with the unphysical filter disabled.
    \item \textbf{Str2Str}~\cite{lundstr2str}: forward-noise level $\Delta t{=}0.2$.
\end{itemize}

For trajectory generation, we use the pretrained ConfRover-Base~\cite{shen2025simultaneous} model, retrain MDGEN~\cite{jing2024generative} on the DynamicPDB training set, and train another generator from scratch with the AlphaFolding~\cite{cheng20254d} architecture, but replacing its OmegaFold sequence embedding with Evoformer. All trajectory-generation methods produce rollouts of approximately $100$\,ns at a $0.4$\,ns sampling interval, yielding $256$ frames per trajectory.

\section{Bias and Refinement Decomposition on Fast-Folding}
\label{app:bias_refine_control}

To disentangle the contributions of the learned bias and the score-based refinement step, we evaluate the full $\eta$ vs.\ refinement grid on the 12 Fast-Folding proteins at $E{=}640$ and $3$ seeds, varying the bias strength $\eta\in\{0,\,0.05,\,0.1,\,0.2,\,0.4\}$ and toggling refinement at $t_{\mathrm{ref}}{=}0.2$. Table~\ref{tab:bias_refine_control} reports peak coverage, time-to-coverage~(TTC), and the corresponding speedup over the baseline.

\begin{table}[h]
\centering
\caption{\textbf{$\eta\times$\{refinement\} control on Fast-Folding.} Speedup is computed against the unbiased $\eta{=}0$ no-refinement reference; ``--'' indicates that the configuration never reaches that reference within the $E{=}640$ budget. Pooled means across $12$ proteins $\times\ 3$ seeds; SEM in parentheses.}
\label{tab:bias_refine_control}
\small
\setlength{\tabcolsep}{6pt}
\begin{tabular}{cccc cccc}
\toprule
 & \multicolumn{3}{c}{No refinement} & & \multicolumn{3}{c}{Refinement at $t_{\mathrm{ref}}{=}0.2$} \\
\cmidrule(lr){2-4} \cmidrule(lr){6-8}
$\eta$ & max coverage $\uparrow$ & TTC $\downarrow$ & speedup $\uparrow$ & & max coverage $\uparrow$ & TTC $\downarrow$ & speedup $\uparrow$ \\
\midrule
$0$    & $0.208\,(0.025)$ & $637$ & $1.0\times$  & & $0.203\,(0.024)$ & $637$ & $1.0\times$  \\
$0.05$ & $0.426\,(0.033)$ & $84$  & $7.6\times$  & & $0.409\,(0.031)$ & $82$  & $7.8\times$  \\
$0.1$  & $0.602\,(0.035)$ & $43$  & $14.8\times$ & & $0.636\,(0.035)$ & $40$  & $15.9\times$ \\
$0.2$  & $0.438\,(0.033)$ & $96$  & $6.6\times$  & & $0.678\,(0.033)$ & $17$ & $37.5\times$ \\
$0.4$  & $0.156\,(0.035)$ & --    & --           & & $0.669\,(0.028)$ & $12$  & $53.1\times$ \\
\bottomrule
\end{tabular}
\end{table}

\section{History-Dependent Implicit Bias}
\label{sec:appendix_tilt}

\subsection{From Enhanced Sampling to Relative Visitation}
\label{subsec:tilt_relative_visitation}

\paragraph{Enhanced Sampling:} Classical enhanced sampling starts from a Boltzmann density
\begin{equation}
p_{\mathrm{eq}}(\mathbf{x})
=Z^{-1}\exp[-\beta U(\mathbf{x})].
\end{equation}
Adding a bias potential $V_{\mathrm{bias}}$ changes the sampled density to
\begin{equation}
p_{\mathrm{bias}}(\mathbf{x})
\propto
p_{\mathrm{eq}}(\mathbf{x})
\exp[-\beta V_{\mathrm{bias}}(\mathbf{x})].
\label{eq:tilt_enhanced_sampling}
\end{equation}
In conventional history-dependent enhanced sampling, such as metadynamics, a positive bias is accumulated in previously visited regions of collective-variable space, thereby discouraging repeated occupation~\cite{laio2002escaping,barducci2008well}. Motivated by this principle, we construct a generative-space bias that penalizes a region only when its historical occupation is excessive relative to the emulator distribution. The corresponding score correction therefore points away from such relatively over-occupied regions.

\paragraph{Estimation of History Density:} Let $p_{\mathrm{env},t}(\mathbf{x}_t)$ denote the marginal density represented by the frozen emulator at time $t$. Fix $t>0$, and let all densities below be defined with respect to a common reference measure on the state space. For a history bank $\mathcal{H}=\{\mathbf{h}^{0},\ldots,\mathbf{h}^{H-1}\}$, define the smoothed history-occupation density by
\begin{equation}
p_{\mathrm{hist},t}(\mathbf{x}_t\mid\mathcal{H})
=
\sum_{j=0}^{H-1}\pi_j
p_t(\mathbf{x}_t\mid\mathbf{h}^{j}),
\qquad
\pi_j\geq0,
\quad
\sum_j\pi_j=1,
\label{eq:tilt_history_mixture}
\end{equation}
where $p_t(\mathbf{x}_t\mid\mathbf{h}^{j})$ is the forward perturbation kernel for the diffusion model. 

\paragraph{Estimation of Visitation:} A high historical occupation does not by itself imply over-visitation: a basin may be visited frequently simply because it has high probability under the emulator. We therefore assess over-visitation relative to the emulator's expected occupation.
\begin{equation}
\rho_{\mathcal{H},t}(\mathbf{x}_t)
:=
\frac{p_{\mathrm{hist},t}(\mathbf{x}_t\mid\mathcal{H})}
{p_{\mathrm{env},t}(\mathbf{x}_t)}.
\label{eq:tilt_relative_occupation}
\end{equation}
Regions with $\rho_{\mathcal{H},t}>1$ and $\rho_{\mathcal{H},t}<1$ are over- and under-occupied relative to the frozen emulator, respectively.

\paragraph{From Relative Visitation to a Bias Potential:} The Boltzmann relationship between population and free energy motivates representing a strictly positive density $p$ by the effective energy $F_p=-\beta^{-1}\log p+C$, defined relative to the chosen reference measure. We therefore map the relative occupation in Eq.~\eqref{eq:tilt_relative_occupation} directly to the dimensionless bias $\beta V_{\mathcal{H},t}$:
\begin{equation}
\beta V_{\mathcal{H},t}(\mathbf{x}_t)
:=
\eta\log\rho_{\mathcal{H},t}(\mathbf{x}_t)
=
\eta\log\frac{p_{\mathrm{hist},t}(\mathbf{x}_t\mid\mathcal{H})}
{p_{\mathrm{env},t}(\mathbf{x}_t)},
\qquad \eta\geq0.
\label{eq:tilt_bias_potential}
\end{equation}
Regions with $\rho_{\mathcal{H},t}>1$ receive a positive bias and are suppressed by $\exp(-\beta V_{\mathcal{H},t})$; under-occupied regions receive a negative bias and are promoted. Thus the logarithm converts relative occupation into an additive bias potential, while $\eta$ controls its strength.

\subsection{The Induced Tilt and Score Bias}
\label{subsec:tilt_score_derivation}

Applying Eq.~\eqref{eq:tilt_enhanced_sampling} to the emulator marginal and substituting Eq.~\eqref{eq:tilt_bias_potential} gives the fixed-time tilted density
\begin{equation}
q_{\eta,t}(\mathbf{x}_t\mid\mathcal{H})
=
\frac{1}{Z_{\eta,t}(\mathcal{H})}
p_{\mathrm{env},t}(\mathbf{x}_t)
\left(
\frac{p_{\mathrm{env},t}(\mathbf{x}_t)}
{p_{\mathrm{hist},t}(\mathbf{x}_t\mid\mathcal{H})}
\right)^{\eta},
\label{eq:tilt_target_appendix}
\end{equation}
Throughout, we assume that $p_{\mathrm{env},t}$ and $p_{\mathrm{hist},t}$ are strictly positive on a common support and that $Z_{\eta,t}(\mathcal{H})<\infty$. This tilt reduces to the frozen emulator at $\eta=0$ and up-weights regions that the emulator supports more strongly than the rollout history occupies.

Define the ideal marginal scores
\begin{equation}
\mathbf{s}_{\mathrm{env}}
:=\nabla_{\mathbf{x}_t}\log p_{\mathrm{env},t},
\qquad
\mathbf{s}_{\mathrm{hist}}
:=\nabla_{\mathbf{x}_t}\log p_{\mathrm{hist},t}.
\end{equation}
Here $\mathbf{s}_{\mathrm{env}}$ denotes the exact emulator-marginal score; in implementation, it is approximated by the frozen score network $\mathbf{s}_{\mathrm{emu}}^\theta$ of Section~\ref{subsec:bias_score}.
Differentiating the log-density in Eq.~\eqref{eq:tilt_target_appendix} and using the fact that $Z_{\eta,t}$ is independent of $\mathbf{x}_t$ yields
\begin{align}
\mathbf{s}_{\pi}^{\star}
&:=\nabla_{\mathbf{x}_t}\log q_{\eta,t}
\nonumber\\
&=(1+\eta)\mathbf{s}_{\mathrm{env}}
-\eta\mathbf{s}_{\mathrm{hist}}
\nonumber\\
&=\mathbf{s}_{\mathrm{env}}
+\eta\underbrace{(
\mathbf{s}_{\mathrm{env}}-\mathbf{s}_{\mathrm{hist}}
)}_{\mathbf{s}_{\mathrm{bias}}^{\star}}.
\label{eq:tilt_score_identity}
\end{align}
Equivalently,
\begin{equation}
\mathbf{s}_{\mathrm{bias}}^{\star}
=\nabla_{\mathbf{x}_t}
\log\frac{p_{\mathrm{env},t}}
{p_{\mathrm{hist},t}},
\qquad
-\beta\nabla_{\mathbf{x}_t}V_{\mathcal{H},t}
=\eta\mathbf{s}_{\mathrm{bias}}^{\star}.
\label{eq:tilt_bias_force}
\end{equation}
Hence the density-ratio score is the generative-space counterpart of the force induced by the history-dependent bias potential: it subtracts the score component associated with occupied history while retaining the emulator score as a reference.

Equation~\eqref{eq:tilt_target_appendix} is defined directly at each $t$. In general, forward noising does not commute with products, powers, or ratios of densities, so the family $\{q_{\eta,t}\}_t$ need not be the noised marginals of a single clean-space power-law tilt. Our claim is the fixed-time score identity in Eq.~\eqref{eq:tilt_score_identity}, which is the identity used to construct the reverse score field.

\subsection{History Mixture and Branchwise Estimation}
\label{subsec:tilt_history_branches}
To evaluate the right-hand side of Eq.~\eqref{eq:tilt_score_identity}, we need to estimate the history score $\mathbf{s}_{\mathrm{hist}}$. We use the history marginal of Eq.~\eqref{eq:tilt_history_mixture}, a mixture of forward diffusion kernels $p_t(\mathbf{x}_t\mid\mathbf{h}^{j})$ with weights $\{\pi_j\}$ independent of $\mathbf{x}_t$. The history score is
\begin{align}
\mathbf{s}_{\mathrm{hist}}
&=
\nabla_{\mathbf{x}_t}
\log p_{\mathrm{hist},t}(\mathbf{x}_t\mid\mathcal{H})
\nonumber\\
&=
\nabla_{\mathbf{x}_t}
\log
\sum_j
\pi_j p_t(\mathbf{x}_t\mid\mathbf{h}^{j})
\nonumber\\
&=
\frac{
\sum_j\pi_j
\nabla_{\mathbf{x}_t}
p_t(\mathbf{x}_t\mid\mathbf{h}^{j})
}{
\sum_l\pi_l
p_t(\mathbf{x}_t\mid\mathbf{h}^{l})
}
\nonumber\\
&=
\sum_j
\underbrace{
\frac{
\pi_j\,p_t(\mathbf{x}_t\mid\mathbf{h}^{j})
}{
\sum_l\pi_l\,p_t(\mathbf{x}_t\mid\mathbf{h}^{l})
}
}_{\Gamma_j(\mathbf{x}_t,t,\mathcal{H})}
\nabla_{\mathbf{x}_t}
\log p_t(\mathbf{x}_t\mid\mathbf{h}^{j})
\nonumber\\
&=
\sum_j
\Gamma_j
\mathbf{s}_{\mathrm{hist},j}^{\star}.
\label{eq:tilt_history_score}
\end{align}
Here $\Gamma_j$ is the posterior responsibility of history branch $j$, and $\mathbf{s}_{\mathrm{hist},j}^{\star}$ is the known forward kernel score. Because $\sum_j\Gamma_j=1$, the ideal bias in Eq.~\eqref{eq:tilt_score_identity} admits the branchwise decomposition
\begin{equation}
\mathbf{s}_{\mathrm{bias}}^{\star}
=
\mathbf{s}_{\mathrm{env}}-\sum_{j}\Gamma_j\mathbf{s}_{\mathrm{hist},j}^{\star}
=
\sum_{j}\Gamma_j\bigl(\mathbf{s}_{\mathrm{env}}-\mathbf{s}_{\mathrm{hist},j}^{\star}\bigr),
\label{eq:tilt_bias_branchwise}
\end{equation}
which has exactly the structural form of the branchwise estimator in Eq.~\eqref{eq:effective_bias_score}. We therefore approximate this ideal bias with that estimator and train each branch with the tractable per-branch objective in Eq.~\eqref{eq:diff_loss}.

Because the forward perturbation kernel is known, each ideal component score $\mathbf{s}_{\mathrm{hist},j}^{\star}=\nabla_{\mathbf{x}_t}\log p_t(\mathbf{x}_t\mid\mathbf{h}^{j})$ is available in closed form. Equation~\eqref{eq:diff_loss} supervises each history branch toward its corresponding component score. We use the RMSD-based weights in Eq.~\eqref{eq:history_branch_weight} as a heuristic proxy for the ideal mixture responsibilities $\Gamma_j$ in Eq.~\eqref{eq:tilt_history_score}.

\subsection{Variational Characterization}
\label{subsec:tilt_variational}

The fixed-time tilt in Eq.~\eqref{eq:tilt_target_appendix} admits an equivalent variational characterization. Define the relative exploration reward
\begin{equation}
R_{\mathcal{H},t}(\mathbf{x}_t)
:=
\log\frac{p_{\mathrm{env},t}(\mathbf{x}_t)}
{p_{\mathrm{hist},t}(\mathbf{x}_t\mid\mathcal{H})},
\label{eq:tilt_relative_reward}
\end{equation}
and let $\mathcal{P}_t$ denote the set of normalized densities that are absolutely continuous with respect to $p_{\mathrm{env},t}$ and for which the expectation and KL divergence below are finite. For $q\in\mathcal{P}_t$, consider the KL-regularized functional
\begin{equation}
\mathcal{F}_t(q)
:=
\eta\,\mathbb{E}_{\mathbf{x}_t\sim q}
\bigl[R_{\mathcal{H},t}(\mathbf{x}_t)\bigr]
-D_{\mathrm{KL}}(q\,\|\,p_{\mathrm{env},t}).
\label{eq:tilt_variational_objective}
\end{equation}
The reward term favors states whose historical occupation is small relative to their emulator probability. The reverse-KL term regularizes this reallocation of probability mass toward the emulator reference; without it, maximizing the linear reward functional may concentrate $q$ on the essential maximizers of $R_{\mathcal{H},t}$.

Introducing a Lagrange multiplier $\zeta$ for $\int q(\mathbf{x}_t)\,d\mathbf{x}_t=1$ gives
\begin{align}
\mathcal{J}_t[q,\zeta]
&=
\eta\int q(\mathbf{x}_t)R_{\mathcal{H},t}(\mathbf{x}_t)\,d\mathbf{x}_t
\nonumber\\
&\quad-
\int q(\mathbf{x}_t)
\log\frac{q(\mathbf{x}_t)}{p_{\mathrm{env},t}(\mathbf{x}_t)}
\,d\mathbf{x}_t
\nonumber\\
&\quad+
\zeta\left(\int q(\mathbf{x}_t)\,d\mathbf{x}_t-1\right).
\label{eq:tilt_variational_lagrangian}
\end{align}
Its first variation is
\begin{equation}
\frac{\delta\mathcal{J}_t}{\delta q(\mathbf{x}_t)}
=
\eta R_{\mathcal{H},t}(\mathbf{x}_t)
-\log\frac{q(\mathbf{x}_t)}{p_{\mathrm{env},t}(\mathbf{x}_t)}
-1+\zeta.
\label{eq:tilt_first_variation}
\end{equation}
The stationarity and normalization conditions yield
\begin{equation}
q_t^{\star}(\mathbf{x}_t)
=
\frac{1}{Z_{\eta,t}}
p_{\mathrm{env},t}(\mathbf{x}_t)
\exp\!\left[\eta R_{\mathcal{H},t}(\mathbf{x}_t)\right],
\qquad
Z_{\eta,t}
=
\int p_{\mathrm{env},t}(\mathbf{x}_t)
\exp\!\left[\eta R_{\mathcal{H},t}(\mathbf{x}_t)\right]
\,d\mathbf{x}_t.
\label{eq:tilt_variational_solution}
\end{equation}
The functional in Eq.~\eqref{eq:tilt_variational_objective} is the sum of a linear term and the negative KL, which is strictly concave in $q$; hence this stationary density is the unique maximizer whenever $Z_{\eta,t}<\infty$. Substitution of Eq.~\eqref{eq:tilt_relative_reward} into Eq.~\eqref{eq:tilt_variational_solution} recovers Eq.~\eqref{eq:tilt_target_appendix}.

\section{Pathwise Measure Interpretation}
\label{sec:appendix_pathwise}

\subsection{Trajectory Measurement}
\label{subsec:trajectory_measures}

To formalize the pathwise measure view, we lift stepwise transition kernels to path measures over reverse denoising trajectories. Consider one reverse denoising trajectory $\xi := \{\mathbf{x}_t\}_{t\in[0,1]}$ generating a single frame of $\tau$, with the trajectory index $k$ suppressed. The reverse process with transition kernel $\mathbb{K}$ and score field $\mathbf{s}$ induces a path measure on trajectory space. In discrete time, its trajectory density factorizes as
\begin{equation}
    p(\xi)
    =
    p(\mathbf{x}_{1})
    \prod_{t}
    \mathbb{K}(\mathbf{x}_{t-\Delta t} \mid \mathbf{x}_{t}, \mathbf{s}),
    \label{eq:trajectory_density_factorization}
\end{equation}
where $p(\mathbf{x}_{1})$ is the prior at $t=1$. We instantiate this factorization for two reverse processes: the frozen emulator process ($\mathbb{K}_{\mathrm{emu}}$ with score $\mathbf{s}_{\mathrm{emu}}^\theta$) and the biased process ($\mathbb{K}_{\pi}$ with biased score); their induced path measures are compared in Appendix~\ref{subsec:girsanov}.

\subsection{Continuous-Time Form of the Support Regularizer}
\label{subsec:girsanov}

The biased reverse score $\mathbf{s}_\pi^\phi$ from Eq.~\eqref{eq:score_perturbation} induces a pathwise change of measure relative to the frozen emulator dynamics. Viewing reverse sampling in continuous time, the frozen emulator model defines the reverse SDE
\begin{equation}
d\mathbf{x}_t = \Big[f(\mathbf{x}_t, t) - g(t)^2\mathbf{s}_{\mathrm{emu}}^\theta(\mathbf{x}_t, t)\Big]dt + g(t)\,d\mathbf{w}_t,
\label{eq:env_sde}
\end{equation}
while substituting the biased reverse score yields
\begin{equation}
d\mathbf{x}_t = \Big[f(\mathbf{x}_t, t) - g(t)^2\mathbf{s}_{\pi}^{\phi}(\mathbf{x}_t, t, \mathcal{H}, \mathbf{c})\Big]dt + g(t)\,d\tilde{\mathbf{w}}_t,
\label{eq:guided_sde}
\end{equation}
where $f(\mathbf{x}_t, t)$ and $g(t)$ are the drift and diffusion coefficients inherited from the forward noise schedule~\cite{song2020score}, and $\tilde{\mathbf{w}}_t$ is a standard Wiener process under the biased path measure $\mathbb{Q}_\pi$. The two SDEs share the diffusion coefficient $g(t)$ and differ only in drift by $-\eta\,g(t)^2\,\mathbf{s}_{\mathrm{bias}}^{\phi}$, equivalently by $g(t)\,\mathbf{u}_t$ with
\begin{equation}
\mathbf{u}_t := -\eta\,g(t)\,\mathbf{s}_{\mathrm{bias}}^{\phi}(\mathbf{x}_t, t, \mathcal{H}, \mathbf{c}),
\label{eq:drift_perturbation_u}
\end{equation}
which identifies the learned bias as a history-dependent drift perturbation of the frozen reverse dynamics.

Denote by $\mathbb{P}_{\mathrm{emu}}$ and $\mathbb{Q}_\pi$ the path measures induced by Eq.~\eqref{eq:env_sde} and Eq.~\eqref{eq:guided_sde}; their log-likelihood ratio admits a continuous It\^{o} form and an equivalent discrete per-step form. In continuous time, Girsanov's theorem~\cite{song2020score} gives the ratio as an It\^{o} integral plus a quadratic-variation correction,
\begin{equation}
\log \frac{d\mathbb{Q}_\pi}{d\mathbb{P}_{\mathrm{emu}}}
=
\int_{T}^{0} \mathbf{u}_t \cdot d\mathbf{w}_t
\;-\;
\frac{1}{2}\int_{T}^{0} \|\mathbf{u}_t\|^2\,dt,
\label{eq:girsanov_ito}
\end{equation}
with $\mathbf{u}_t$ the normalized drift perturbation from Eq.~\eqref{eq:drift_perturbation_u}. The quadratic term is a free-energy-like accumulation of squared bias magnitude along the path. Under Euler--Maruyama discretization, the Radon--Nikodym derivative factorizes over reverse steps into a product of per-step kernel ratios,
\begin{equation}
\frac{d\mathbb{Q}_\pi}{d\mathbb{P}_{\mathrm{emu}}}(\xi)
=
\prod_t
\frac{\mathbb{K}_\pi(\mathbf{x}_{t-\Delta t}\mid \mathbf{x}_t, \mathcal{H}, t)}
     {\mathbb{K}_{\mathrm{emu}}(\mathbf{x}_{t-\Delta t}\mid \mathbf{x}_t, \mathbf{s}_{\mathrm{emu}}^\theta(\mathbf{x}_t, t))},
\label{eq:rn_discrete_product}
\end{equation}
where $\xi=\{\mathbf{x}_t\}_{t\in[0,1]}$ denotes a reverse denoising path, and $d\mathbb{Q}_\pi/d\mathbb{P}_{\mathrm{emu}}$ is the Radon--Nikodym density of the biased path measure relative to the emulator path measure. Taking the logarithm decomposes the pathwise log-likelihood ratio into a sum of step-wise log-density ratios,
\begin{equation}
\log \frac{d\mathbb{Q}_\pi}{d\mathbb{P}_{\mathrm{emu}}}(\xi)
=
\sum_t
\underbrace{
\log
\frac{\mathbb{K}_\pi(\mathbf{x}_{t-\Delta t}\mid \mathbf{x}_t, \mathcal{H}, t)}
     {\mathbb{K}_{\mathrm{emu}}(\mathbf{x}_{t-\Delta t}\mid \mathbf{x}_t, \mathbf{s}_{\mathrm{emu}}^\theta(\mathbf{x}_t, t))}
}_{\Delta_b^t}.
\label{eq:rn_equals_delta_sum}
\end{equation}

Equations~\eqref{eq:girsanov_ito} and~\eqref{eq:rn_equals_delta_sum} together place the learned bias in a continuous-time SDE form: the discrete sum $\sum_t \Delta_b^t$ relates expectations under $\mathbb{P}_{\mathrm{emu}}$ to those under $\mathbb{Q}_\pi$. The integrand of the environment-support regularizer in Eq.~\eqref{eq:support_loss_frame} coincides with $\Delta_b^t$ evaluated at a single Monte Carlo sample, so $\mathcal{L}_{\mathrm{support}}$ matches the per-step integrand of the pathwise log-density ratio.

\end{document}